\documentclass{article}
\usepackage{jfrExamplee}
\usepackage{graphicx}
\usepackage{apalike}
\usepackage{setspace}
\usepackage{todonotes}
\usepackage{url}
\usepackage{array, makecell}
\usepackage{multirow}
\usepackage{amsmath}
\usepackage[caption=false]{subfig}
\usepackage{soul}

\title{Autonomous, Mobile Manipulation in a Wall-building Scenario: Team LARICS at MBZIRC 2020}

\author{
Ivo Vatavuk, Marsela Polic, Ivan Hrabar, Frano Petric, Matko Orsag, Stjepan Bogdan\\
Laboratory for Robotics and Intelligent Control systems\\
University of Zagreb Faculty of Electrical Engineering and Computing\\
Unska 3, Zagreb, Croatia \\
\texttt{ivo.vatavuk@fer.hr, larics@fer.hr}
}

\begin{document}

\maketitle

\begin{abstract}

In this paper we present our hardware design and control approaches for a mobile manipulation platform used in Challenge 2 of the MBZIRC 2020 competition.
In this challenge, a team of UAVs and a single UGV collaborate in an autonomous, wall-building scenario, motivated by construction automation and large-scale robotic 3D printing. The robots must be able, autonomously, to detect, manipulate, and transport bricks in an unstructured, outdoor environment. Our control approach is based on a state machine that dictates which controllers are active at each stage of the Challenge.
In the first stage our UGV uses visual servoing and local controllers to approach the target object without considering its orientation. The second stage consists of detecting the object's global pose using OpenCV-based processing of RGB-D image and point-cloud data, and calculating an alignment goal within a global map. The map is built with Google Cartographer and is based on onboard LIDAR, IMU, and GPS data. Motion control in the second stage is realized using the ROS Move Base package with Time-Elastic Band trajectory optimization. Visual servo algorithms guide the vehicle in local object-approach movement and the arm in manipulating bricks. To ensure a stable grasp of the brick's magnetic patch, we developed a passively-compliant, electromagnetic gripper with tactile feedback.
Our fully-autonomous UGV performed well in Challenge 2 and in post-competition evaluations of its brick pick-and-place algorithms.

\end{abstract}

\section{Introduction}\label{sec:intro}

The focus of many robotics competitions in recent years has been on tasks closely related to warehouse automation or service robotics. Such competitions include the Amazon Picking Challenge \cite{Corell2018} and the RoboCup@Home challenge \cite{Matamoros2019}, which both emphasize object-detection and grasp-planning for pick-and-place tasks. The Mohamed Bin Zayed International Robotics Challenge \cite{mbzirc} builds upon these ideas and extends the complexity by placing the robots outdoors in a relatively unstructured environment. For the 2020 edition of MBZIRC, one of the challenges emulates a construction site where a heterogeneous multi-robot team, consisting of multiple, unmanned, aerial vehicles (UAVs) and a single, unmanned, ground vehicle (UGV) must build a wall with a predefined blueprint.

The wall-building scenario can be easily generalized to other real-world problems involving navigation, object perception, motion, grasp-planning and task-planning. Warehouse automation and search-and-rescue missions are a few among a plethora of such everyday work environments that have not yet been automated. As with all robotics competitions, the challenges presented should be viewed as a validation of robotic progress and its adaptability to unstructured environments with unmodeled disturbances, and outside of laboratory conditions. The objective of these competitions is to provide a challenge to the robotics research community, to test and validate the existing state of the art, to identify problem areas, and to accelerate scientific progress. No less importantly, the competition can be seen as an opportunity for the less-prominent research groups to gain visibility.

\begin{figure} [htb]
	\centering
	\includegraphics[width=1.0\textwidth]{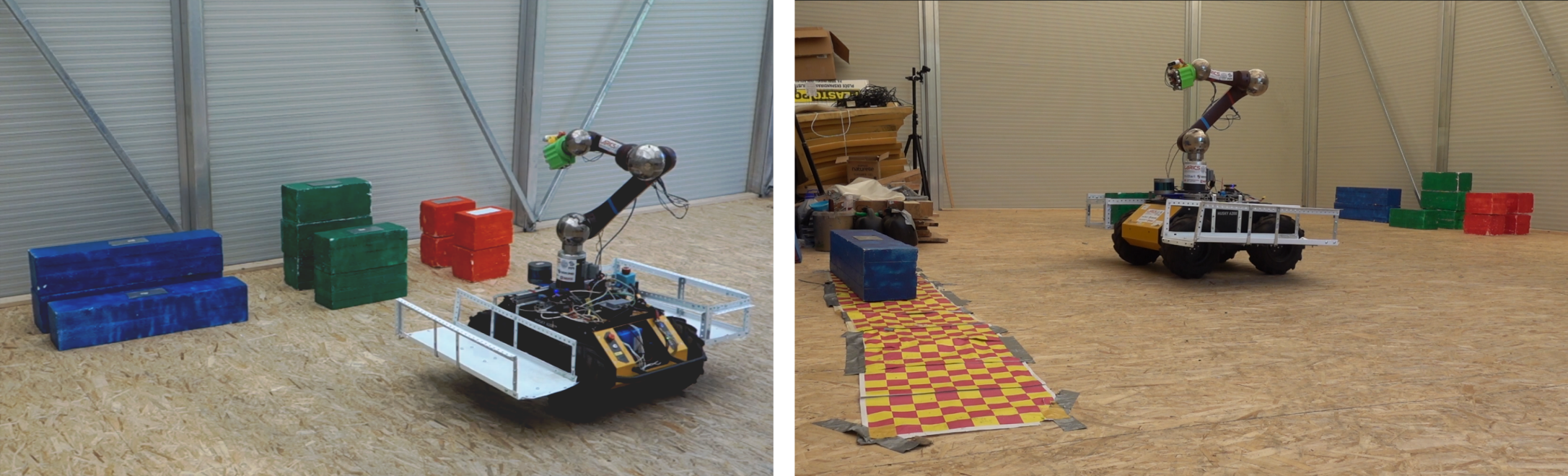}
	\caption{MBZIRC Challenge 2 scenario: A UGV must autonomously find a stack of bricks (left), select and pick up a brick, transport it to a building site marked with a checkered pattern, and place it precisely on the building location (right).}
	\label{fig:husky_bricks}
\end{figure}

Heterogeneous multi-robot teams constitute a well-researched concept with a focus on coordination and planning, as discussed in our earlier paper \cite{Krizmancic2020}. Herein, the focus is on the problems faced by the UGV in the Challenge 2 of MBZIRC 2020. As stated in the challenge rules, the UGV must find the stacks of bricks, which consist of bricks of different colors, transport them to the building site marked by a distinctive pattern on the ground (see Figure \ref{fig:husky_bricks}), and build a wall with a specified brick arrangement. Points are awarded for the correct arrangement and precision of bricklaying and, most importantly, for autonomous task execution.

To complete Challenge 2 the UGV must solve several, nested problems. First, the UGV must navigate precisely in an outdoor environment, finding key locations, transporting bricks, and avoiding brick stacks and any already-constructed wall. In addition, the UGV must autonomously detect the brick stack, a single, target brick atop the stack, and, finally the construction location. Once these tasks are handled, the UGV must accurately determine the grasping point to pick up a brick and then move to position itself in a pose that makes the brick reachable. Similar tasks are required for the brick-laying part of the Challenge. The robot must accurately determine the location of building site and navigate and position itself to enable successful and precise brick placement. 

This paper discusses our contributions in each of the aforementioned tasks. For precise navigation in an outdoor environment, building on our previous research \cite{vasiljevic:16}, we used a highly-precise graph-SLAM navigation and mapping algorithm based on Google Cartographer \cite{Hess2016}. The algorithm combines 3D data from LIDAR, IMU, and GPS data to build a 3D-map of the environment, which is then projected into a 2D costmap for faster path-planning. Visual and depth data from an RGB-D camera is used to autonomously identify the objects of interest and their positions within the UGV map. The geometry of the objects and properties of the Challenge scenario are used to estimate precisely navigation waypoints that ensure reachability of bricks and building site. Finally, to ensure the robustness of picking up and placing of the bricks, an eye-in-hand visual servo approach is deployed, for which a gripper was designed with an integrated  depth camera, tactile feedback, and passive-compliance components to compensate for uneven surfaces and small differences in the orientation of the UGV with respect to the bricks. The approach to solving the technical challenges posed by the equipment used is also discussed.

The remainder of this paper is organized as follows: Section \ref{sec:related_work} positions our approach with respect to the state-of-the-art in the field. Sections \ref{sec:stateMachine}, \ref{sec:detection} and \ref{sec:low_level_control} present a description of the presented developed algorithms for high-level control based on a state machine, object detection and low-level control for navigation and manipulation. Section \ref{sec:experimental} presents the experimental results, testing the performance of brick pick-up and pattern drop-off individually, along with the performance of the autonomous system as a whole, performing the entire Challenge 2 task autonomously. Section \ref{sec:conclusion} concludes the paper with some comments on future work.

\section{Related Work}\label{sec:related_work}

The manipulation of objects using a robotic ground platform is a well-known concept. Pioneering work in the development of an automated brick laying robotic system in building construction was done as early as 1996 \cite{PRITSCHOW1996}. Although robotic automation is widely used in the automotive industry, it is not yet very present in the construction industry. There are many existing solutions in terms of stationary or fixed automation solutions. However, there are also multi-purpose mobile ground robot systems, 
consisting of mobile robot bases and manipulators, enriched with task-specific end effectors and sensors such as \cite{gawel2019}. Fully autonomous robots for building construction are mostly demonstrated under laboratory conditions, but there is a strong industrial need for automation in building construction \cite{DAVILA2019} \cite{MELENBRINK2020}.

Nowadays, object detection is usually performed with neural networks that provide fast and reliable detection, such as Yolo \cite{Bochkovskiy2020} or Fast R-CNN \cite{Girshick2015}. However, taking advantage of the fact that the objects in the Challenge 2 of MBZIRC 2020 are color-coded, the approach in this work is based on classical detection in HSV space, with several extensions for increased reliability, similar to those presented in \cite{Nieuwenhuisen2017}. 

Since the goal of the Challenge is to detect and manipulate objects, the robot must estimate the pose of the brick and evaluate whether the brick is reachable before planning the grasping maneuver. Since the gripper is equipped with an RGB-D camera, the possibility of using depth data to estimate the pose of the brick was also explored. The most obvious approach is to generate a point cloud from depth data and then perform template matching to extract the pose of a brick \cite{Vock2019}\cite{Opromolla2015}. However, this is computationally expensive for dense point clouds generated by depth cameras. Given the noise observed in depth image data in outdoor experiments, and the fact that the variance in depth estimation for one model of Intel Realsense camera grows quadratically with distance \cite{Ahn2019}, brick-pose estimation could not rely on depth information, but rather on image based estimation.

Similar to object detection, the estimation of objects pose from RGB images is most commonly approached with neural networks, as presented in \cite{Grabner2018} and \cite{Xiang2018}. While offering ever-increasing performance, neural network-based solutions still suffer from the lack of annotated data (see \cite{Kundu2018} and \cite{Sock2020}). A priori knowledge that the objects of interest are positioned in a certain way in the environment, and knowledge of the pose of the camera with respect to the same environment, enable solutions that do not require neural networks. The approach of this work to determining the object's pose is to invert a well-known Perspective-N-Point problem \cite{Fischler1981} and extract the position of 3D points knowing the position of a calibrated camera in the environment, enabled by precise robot localization and manipulator kinematics. 

Once the object of interest is detected and its pose is estimated, the goal of the system is to navigate the mobile robot to the vicinity of the object, followed by precise control of the manipulator to execute a given task. This two-stage approach, where the mobile robot and the arm are controlled separately (while the robot moves the arm is stationary and vice-versa), is inspired by the split-range control concept \cite{ReyesLua2019}. For the navigation part, the TEB planner \cite{Rosmann2017} is used in a 2D costmap obtained from a 3D-map of the environment generated by Google Cartographer. This approach was inspired by the work in \cite{Wulf2004} where the authors use 3D range data to generate 2D maps directly. Since the generation of waypoints in the map was based on image processing, i.e. estimation of objects pose in robot's coordinate frame, the navigation of the robot could be compared to position-based visual servoing \cite{Chaumette2006}. However, in our approach only the reference to the controller is provided by the image processing stack, while the feedback is provided by the map localization using Lidar. Control architecture used for both approaching the bricks and the wall, as well as for control of the arm during loading and unloading of the bricks, can be classified as eye-in-hand image-based visual servoing, which is often used in the field \cite{Muslikhin2020}, \cite{Tsay2004}. 
\section{Equipment}\label{sec:equipment}

Our hardware architecture was driven by the MBZIRC experimental setup specifications, which required a robust mobile manipulation platform capable of driving over areas partially covered with pebbles.

\begin{figure*}
	\centering
	\subfloat[Husky A200 with Schunk LWA4P arm and sensors]{
		\includegraphics[height=6cm]{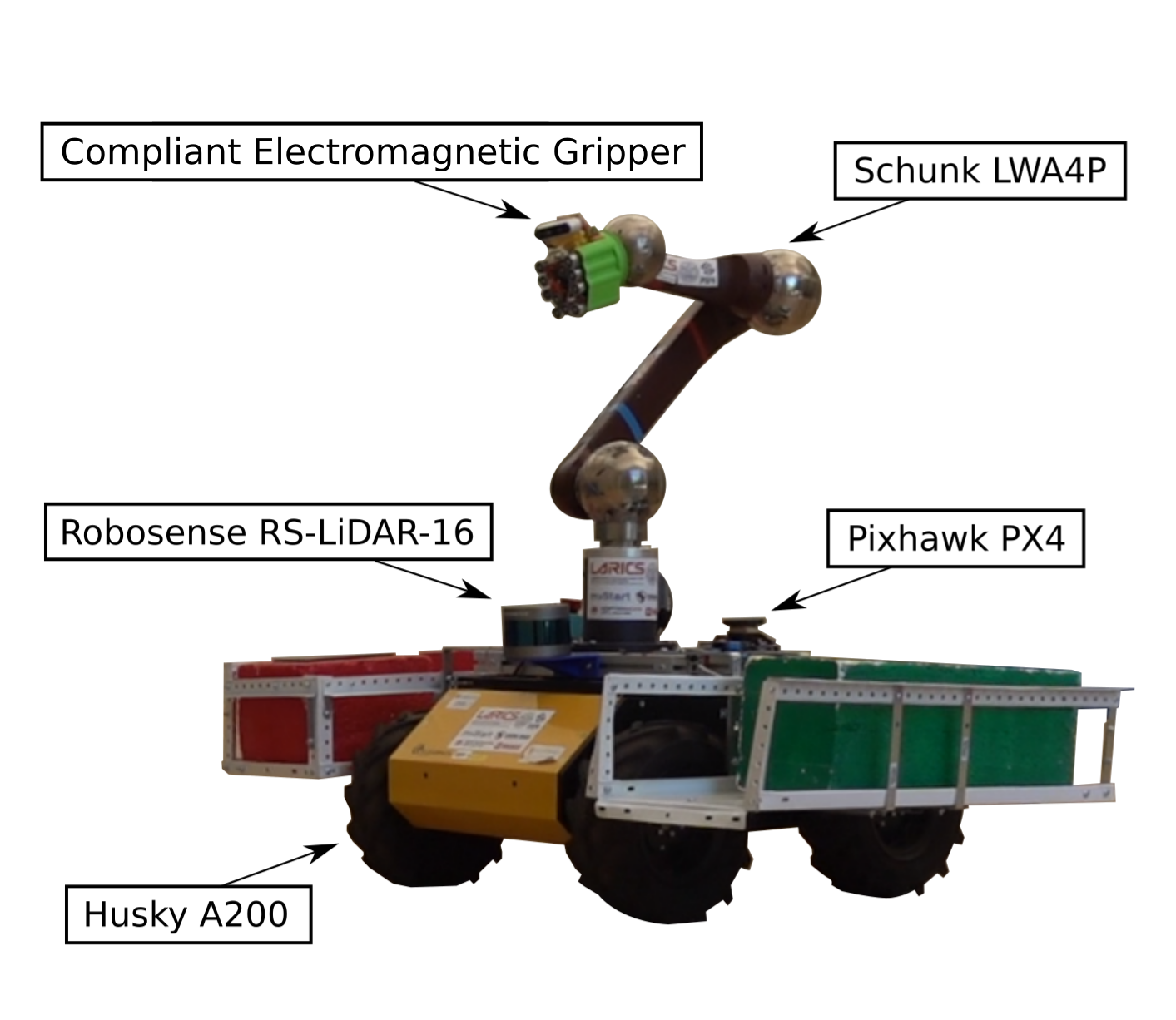}
		\label{fig:platform_a}
	}
	\subfloat[The breakdown of the developed electromagnetic gripper with integrated eye-in-hand depth camera and compliant components which ensure sufficient contact area]{
		\includegraphics[height=6cm]{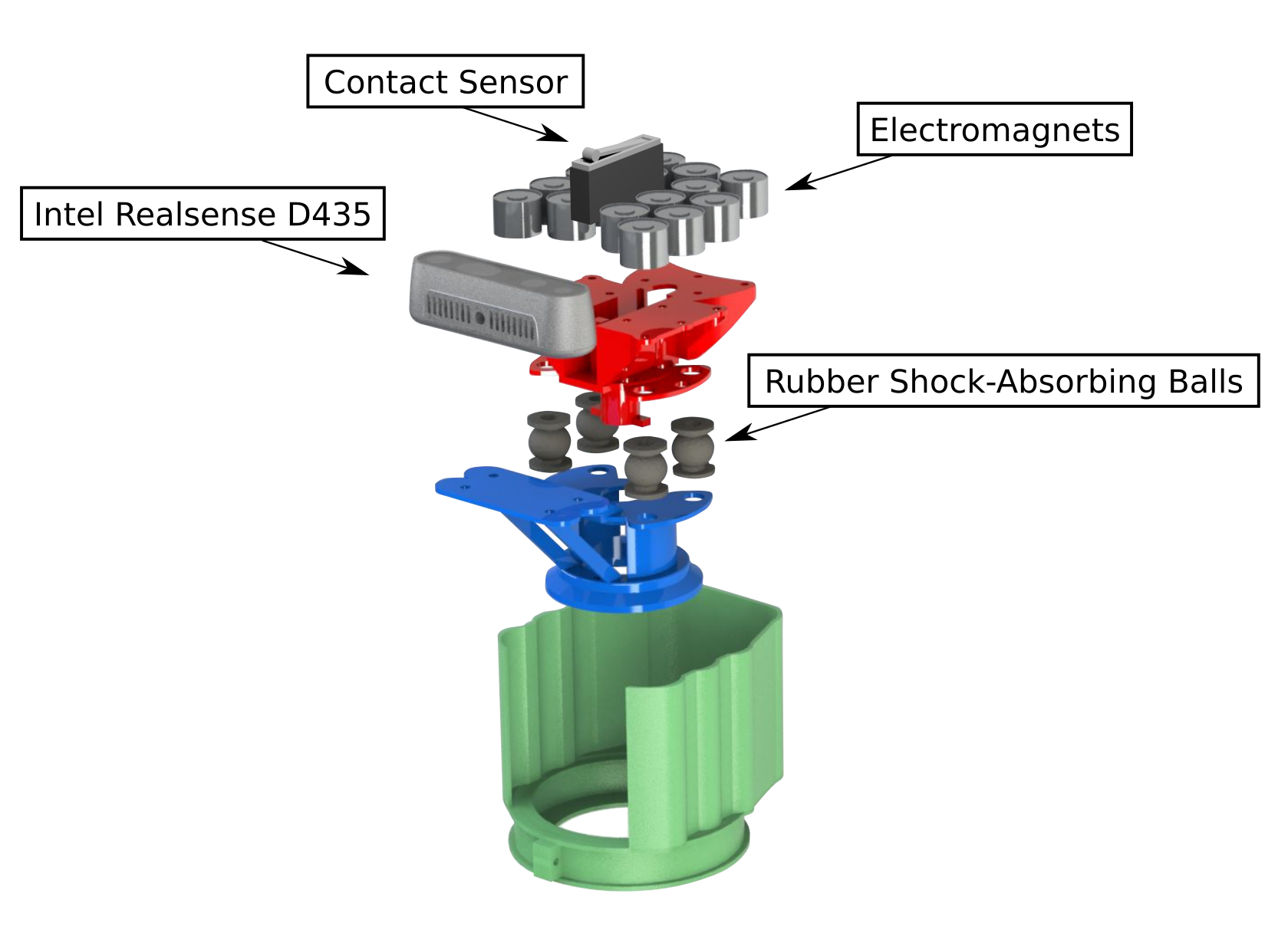}
		\label{fig:platform_b}
	}
	\caption{
		Equipment used by LARICS team in MBZIRC2020. 
	}
	\label{fig:platform}
\end{figure*}

Based on these requirements and our previous experience, we selected the Husky A200, a four-wheeled skid-steered field research platform, manufactured by Clearpath Robotics Inc.
The existing Husky hardware was equipped with an Intel® NUC i7DNHE running Ubuntu 18.04 with ROS Melodic, which served as the primary processing and control computer. Robosense RS-LiDAR-16, a lightweight omnidirectional laser radar with 16 beams, was mounted on the front of the robot (see Figure \ref{fig:platform_a}). To increase mapping accuracy, a Pixhawk autopilot was mounted on the robot, which provided IMU measurements and served as an interface between the Husky on-board control and the remote control unit (RC), which was programmed to immediately switch to manual mode in case of emergency. 

A 6-DoF extended Schunk Powerball LWA4P lightweight arm with integrated joint drives, controlled via a CAN interface, was mounted on the Husky. Having integrated joint drives eliminates the need for additional external control units or power converters. The Schunk manipulator is characterized by its reach of 1460 mm, which is significantly greater than that of other commercially available lightweight arms. To increase the autonomy of the Husky equipped with the Schunk arm, two separate battery packs were installed, with the first primarily powering the robot's drive train, and the second powering the manipulator. In addition to powering a high-current device (the robot drive train or the manipulator), each battery pack powers some low-current electronic components. The output grounds of the battery packs are connected to each other with star-connected grounding cables. To achieve faster turnaround time for placing bricks, additional cargo baskets are installed on both sides of the platform, allowing the Husky to transport up to four bricks, as shown in Figure \ref{fig:platform_a}. 

\subsection{Passively compliant magnetic gripper with integrated depth camera and tactile feedback}\label{sec:gripper}

The bricks in the Challenge 2 of the MBZIRC2020 were designed to have a ferromagnetic area on the top side, which required the design and fabrication of a controllable electromagnetic gripper. Several constraints had to be met in the design of the gripper, which is shown in Figure \ref{fig:platform_b}. Aside from using controllable electromagnets, the gripper had to be lightweight, low-powered and strong enough to lift bricks that can weigh up to 2 kg. 

With weight in mind, ten small electromagnets were selected, each nominally capable of lifting 1 kg. The use of small and low-power magnets in conjunction with a slender ferromagnetic patch on the bricks required that at least six magnets be in contact with the brick to generate magnetic force of sufficient strength to lift the brick. This poses a hard constraint on the orientation of the gripper with respect to the brick when in contact, which could not be satisfied via control algorithms. Due to the stiffness of the Schunk arm, and consequently of the mobile manipulator as a whole, any irregularities on the surface or even slightly different pressures in the robot's tires could cause misalignment of the gripper and the brick, reducing the number of magnets in contact with the ferromagnetic patch. This in turn resulted in the robot being unable to pick up the brick. 

To overcome this problem and ensure full contact of the electromagnetic gripper with the brick at all times, a passive compliance was incorporated into the gripper with four shock-absorbing rubber balls. The rubber balls having clearance between the rigid parts of 15 mm provide the gripper with the ability to compensate for orientation differences of more than $10^\circ$, which ensures optimal contact between the magnets and the surface of the ferromagnetic patch on the brick. The size of the gripper is 10x15cm and it weighs about 700g. 

From the control perspective, laser-based navigation in free space and control based on visual servo principles around the objects of interest are used. An Eye-in-hand approach was implemented, with the Intel Realsense D435 integrated into the gripper, which was used both for detecting objects during navigation and for controlling the mobile manipulator during pickup. To provide the control algorithm with stable references while Husky is stationary, Realsense was mounted on the rigid part of the gripper, right next to the rubber shock absorbers. To mitigate the major drawback of the eye-in-hand visual servo approach, which is loosing feedback when close to the object of interest, a contact sensor in form of a microswitch was added to the gripper, mounted in the center of the end-effector. The role of the tactile feedback was to signal to the visual servo controller that contact had been established, and to provide high-level control with information in case the brick was dropped during transport. 

\section{High-level control} \label{sec:stateMachine}
\begin{figure} [!ht]
	\centering
	\includegraphics[width=0.9\textwidth]{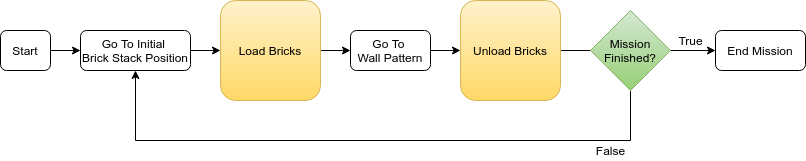}
	\caption{High-level state machine designed for the Challenge. After reaching the brick stack initial position, the desired bricks are loaded into the brick containers. The robot navigates to the wall pattern position, and unloads the bricks, building a portion of the wall. This process is repeated until the mission is completed. }
	\label{fig:sm_1}
\end{figure}

A challenge-specific state machine has been designed that dictates which controllers are used at each stage of the Challenge. In certain states, both the UGV and the manipulator arm must collaborate, while in other states either UGV or manipulator is controlled while the other remains idle. This approach is described using a high-level state machine (Figure \ref{fig:sm_1}) . 

\subsubsection{UGV-only actions}
The mission planner \ref{sec:mission_planning} specifies which bricks to load and unload, providing an input for the high-level state machine (see Figure \ref{fig:planner}). The states \textit{Go To Initial Brick Stack Position} and \textit{Go To Wall Pattern} are \textbf{executed by the UGV}, while the robot arm remains idle. During execution of these two states, the platform navigates through the map to the approximate position of the brick stacks, or the wall pattern, respectively. 

\subsubsection{Collaborative control}
\begin{figure} [!ht]
	\centering
	\includegraphics[width=0.9\textwidth]{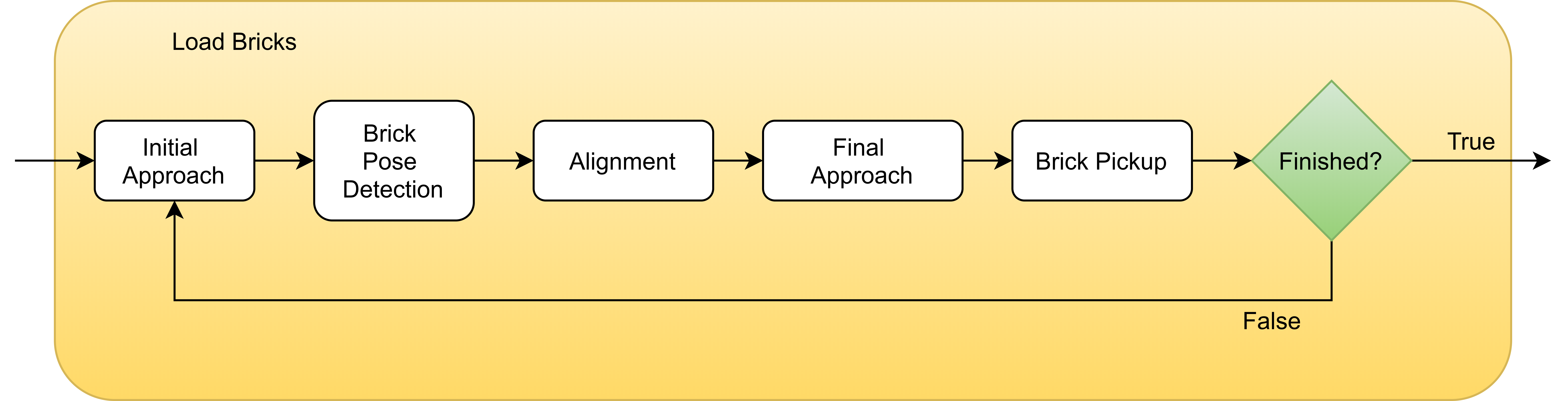}
	\caption{Inside the \textit{Load Bricks} block of the state machine}
	\label{fig:sm_2}
\end{figure}

\begin{figure} [!ht]
	\centering
	\includegraphics[width=1.0\textwidth]{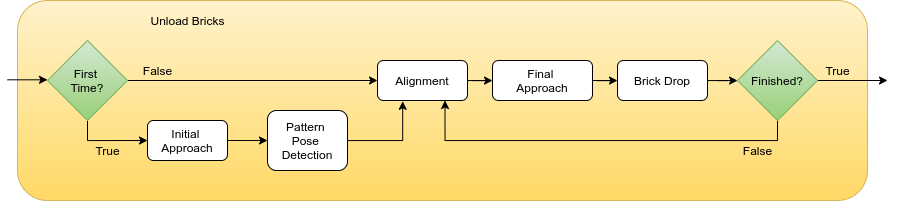}
	\caption{Inside the \textit{Unload Bricks} block of the state machine}
	\label{fig:sm_3}
\end{figure}

Operations in which \textbf{the platform and the robot arm collaborate} are \textit{Load Bricks} and \textit{Unload Bricks}. These two behaviors have similar structures, as shown in Figures \ref{fig:sm_2} and \ref{fig:sm_3}. Here, the platform is controlled by  \textit{Two-Stage Approach}, whose objective is to guide the UGV to a pose where the desired object is within reach of the robot arm. In other words, the emphasis is on the correct orientation of the UGV. Although it is not critical when picking up smaller objects such as the red brick, the correct UGV orientation becomes very important when manipulating larger objects or when approaching the wall pattern.

\begin{figure} [!ht]
	\centering
	\includegraphics[width=0.9\textwidth]{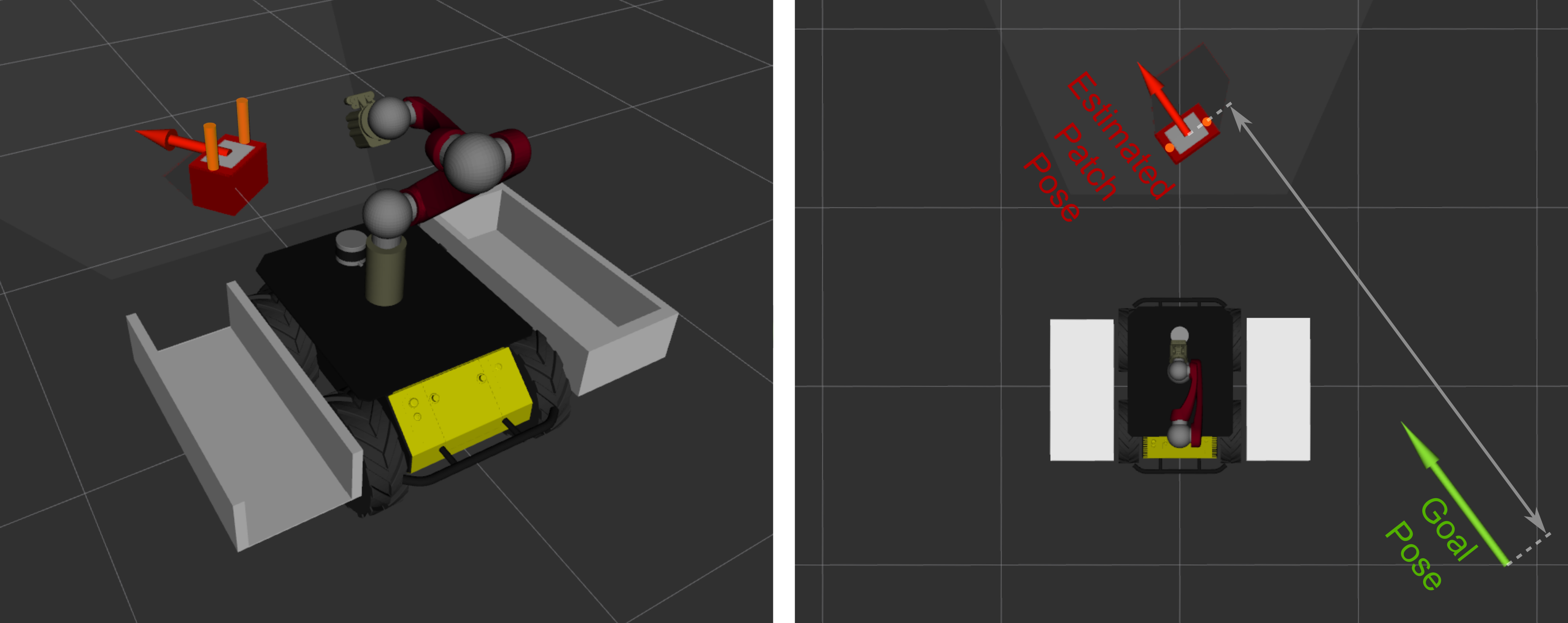}
	\caption{In the \textit{Alignment} state we use target pose detection to calculate an aligned goal for the UGV. The pose of the magnetic patch is obtained in the robot's body frame $L_B$ and transformed into a map frame $L_M$, from which a goal pose is constructed at a fixed distance from the patch with the desired orientation of the robot aligned with the orientation of the brick. This magnetic patch alignment is used in the \textit{Load Bricks} block. A similar principle is employed for the wall footprint alignment in the \textit{Unload Bricks} block.}
	\label{fig:pnp_img}
\end{figure}

Both \textit{Load Bricks} and \textit{Unload Bricks} include \textit{Initial Approach} and \textit{Final Approach} states, in which the robot locally approaches a desired object, without taking its orientation into account. This kind of approach is executed by directly controlling the robot drive and is referred to as \textit{Local Object Approach} in the remainder of this paper. 
\textit{Initial Approach} is used to get close enough to the desired object to detect its pose (\textit{Pose Detection} state), after which a goal pose for the UGV, aligned with the object pose, is calculated (Figure \ref{fig:pnp_img}). Here, the local approach controller is switched off, and the vehicle is stopped. The \textit{Alignment} is conducted with a navigation planner that ensures global orientation of the platform within the map, in close vicinity of the manipulated object. Upon reaching the proper pose, the local approach controller is again switched on. The local control in the \textit{Final Approach} visually guides the platform within the object reach retaining the acquired global orientation. 

The main difference between the \textit{Load Bricks} and \textit{Unload Bricks} behaviors is that the whole two-stage approach is executed only once when unloading bricks onto the wall pattern. All of the subsequent \textit{Unload Bricks} behaviors rely on the pattern's pose saved in the global map frame $L_M$ during the first pose detection state, directly proceeding to the second, \textit{Alignment} stage.

\subsubsection{Manipulator-only actions}
Finally, procedures where\textbf{ only the arm is controlled} while the UGV is idle are \textit{Brick Pickup} and \textit{Brick Drop}. \textit{Brick Pickup} includes running the visual servo algorithm used to grasp the brick with the electromagnetic gripper and placing the brick into an unoccupied brick container. Individual brick placement inside the containers is memorized and used to determine the starting position for the \textit{Brick Drop} visual servo in navigating the arm to the wall pattern.

\subsection{Planning and laying bricks in the correct order}
\label{sec:mission_planning}

\begin{figure} [!htb]
	\centering
	\includegraphics[width=0.9\textwidth]{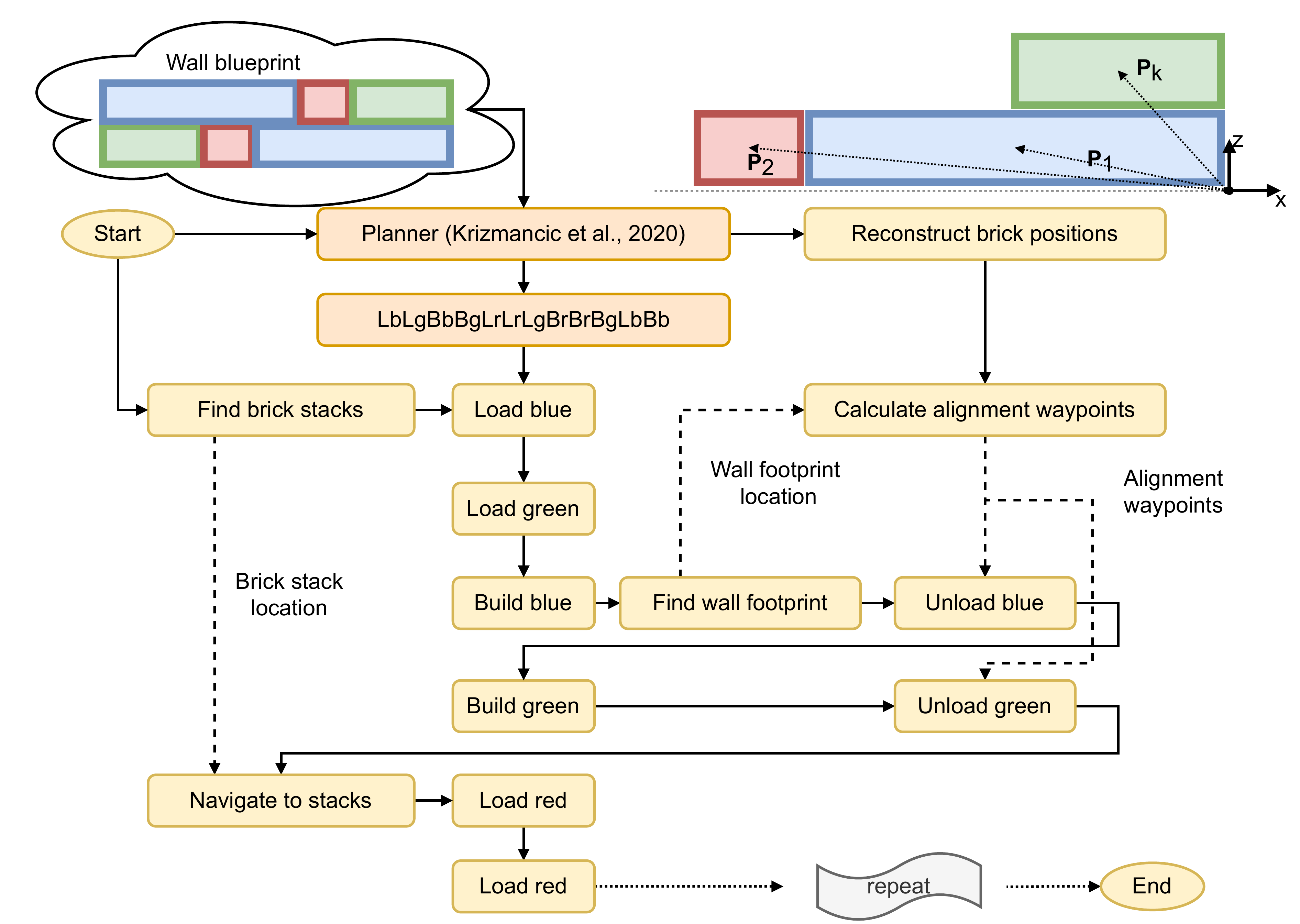}
	\caption{Execution of a wall building scenario. Upon receiving the blueprint, the planner generates a sequence of tasks coded Lc ( where L stands for Load and c is the color of the brick) or Bc (where B stands for build and c is the color of the brick) as the robot searches for bricks (or drives towards bricks if the location is provided by the UAV). Task Bc can first appear after several Lc tasks which can be executed several times until capacity of the robot is filled. In the first building task, the robot searches for the wall footprint (or the location can be provided externally by UAV), while in subsequent runs it uses known brick positions within the blueprint and calculates waypoints based on the known location of the wall footprint and the dimensions of the bricks. Solid arrows show the sequence of actions, while dashed arrows represent non-sequential information flow.}
	\label{fig:planner}
\end{figure}

The blueprint for the wall provided by the referees is parsed by a planner discussed in an earlier paper \cite{Krizmancic2020}, and the optimal sequence of bricks is calculated, taking into account the payload capacity of the robot. 
Algorithm that generates the optimal brick laying order relies on the hierarchical task representation through T\AE MS (Task Analysis, Environment Modeling, and Simulation) language \cite{taems1999}\cite{Lesser2004} and uses the Generalized Partial Global Planning (GPGP) coordination framework \cite{decker1995} to find the optimal sequence of tasks with respect to the robot's capabilities and user-specific cost and/or rewards, which can be specified to minimize the distance traveled, energy expended or in this case, to maximize the number of points in the competition.
The mission for the robot is specified from the optimal brick laying order as shown in Figure \ref{fig:planner}.

Figure \ref{fig:planner} also shows how the state machines from Figures \ref{fig:sm_1}, \ref{fig:sm_2} and \ref{fig:sm_3} integrate to form a complex behavior of the robot. As can be seen, upon receiving the blueprint, the planner generates a sequence of actions (load brick, build brick), that are executed according to the aforementioned state machines. It can be observed that both the brick stacks and the wall footprint are searched for only once, and their location in the global map is used in subsequent trips to the brick stacks and to calculate alignment poses for brick unloading. The alignment poses for unloading are calculated from the now-known pose of the rightmost corner of the wall footprint and the brick positions within the wall blueprint, taking into account a priori known brick dimensions. 

This approach is facilitated by the underlying mapping and localization stack based on the Cartographer, for which we have shown that localization error and drift are minimal (less than 25 cm) due to loop-closing capabilities of Cartographer, even for a more dynamic robot (UAV), large maps (250 m $\times$ 100 m) and large distances covered (more than 600 m), both of which are significantly larger than the arena designated for the ground vehicle in Challenge 2 of MBZIRC2020. For more details on this evaluation, interested reader is invited to read \cite{milijas2020comprehensive}.

\section{Detection of objects of interest}\label{sec:detection}
For \textit{Local Object Approach}, a visual servo algorithm is used to keep observing an object while approaching it. 
This requires detecting the position of the wall footprint pattern, brick stacks, and magnetic patches in an image captured by a camera.
Magnetic patch-detection is also used for the visual servo manipulation algorithm. Here, along with its position, the orientation is also controlled. For the \textit{Two-Stage Approach}, the global pose of the approached object in map frame $L_M$ must be estimated in order to generate an aligned goal pose for the mobile base.
\begin{figure}[htp]
	\centering
	\includegraphics[width=0.5\textwidth]{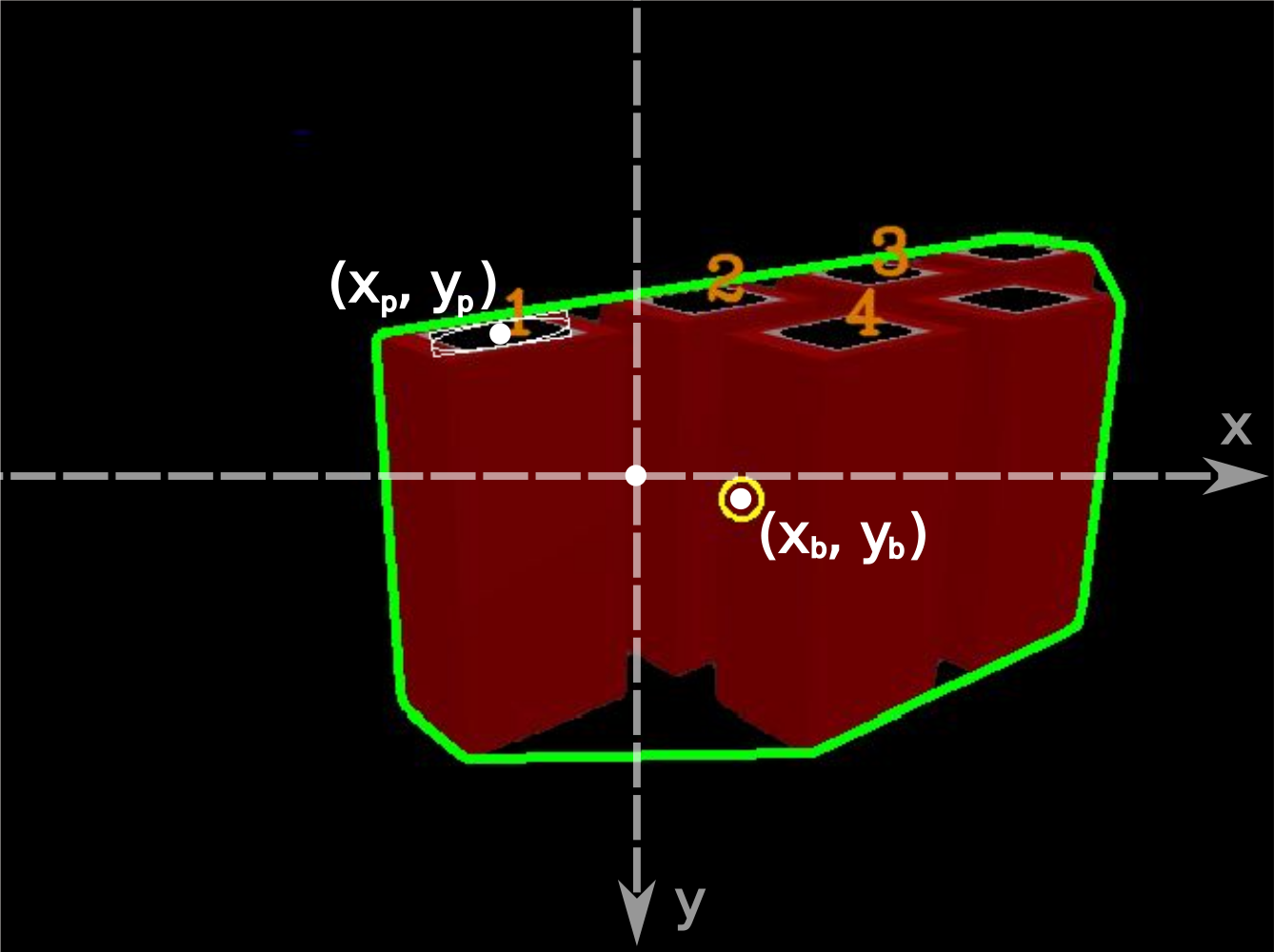}
	\caption{Convex hull of the selected brick stack contour is shown with a green line. The image position of the brick stack is denoted as $(x_b, y_b)$. IDs are assigned to the detected magnetic patches and their positions are tracked from frame to frame. In this particular case, patch number $1$ is selected for approach by using a scoring scheme that takes into account the area of the patch and the patch position in the image (denoted as $(x_p, y_p)$).}
	\label{fig:detekcija}
\end{figure}

\subsection{Brick Stack and Wall Footprint Detection}
To detect the brick stacks, the camera image is filtered using different Hue-Saturation-Value thresholds for red, green, blue and orange bricks. Contour detection is performed on the filtered image, and contours with a sufficiently large surface area are selected as brick stack candidates. As a measure of the quality of a brick stack candidate, it's surface area is used, the ones with a bigger area being of higher quality. The center of the selected brick stack contour is used as its image position ($x_b$, $y_b$), which is used for the local brick stack approach (Figure \ref{fig:detekcija}). 

Wall footprint detection is performed in a similar way, combining Hue-Saturation-Value bounds for yellow and magenta. To start unloading bricks onto the wall footprint, one of its edges must be reached first, and the image position of the rightmost contour point is chosen
as the goal for the \textit{Local Object Approach} corresponding to the upper right corner of the wall footprint. If the rightmost edge is not visible in the image, the camera rotates until the edge comes into view. 

\subsection{Detecting the magnetic patch and calculating its pose}

Contour detection is performed inside the convex hull of a selected brick stack, and detected contours that resemble a rectangle are selected as magnetic patch candidates. The center of the detected rectangle is used as its position in the image ($x_p$, $y_p$) (Figure \ref{fig:detekcija}). 

Since the magnetic patch candidate with the largest area is not necessarily the one the robot must pursue, a scoring scheme is used to calculate the candidate's desirability $S_i$ based on its area and image position. To minimize the rotation required to approach the object, this scoring scheme prefers patch candidates positioned in the center of the image on the $x$ axis. Since objects detected in the lower part of the image are usually closer to the robot, the scoring scheme also favors patches positioned as low as possible on the $y$ axis. The aforementioned logic yields the following equation to calculate the score of the $i$-th patch candidate:
\begin{equation}
S_i = -w_{x} \vert x_{p,i} \vert +
w_{y} y_{p,i} +
w_{A} A_i
\label{eq:score}
\end{equation}
where $A_i$ is the area, $w_{x}$, $w_{y}$ and $w_{A}$ are non-negative weights used to tune the scoring scheme and correspond to the $x$ and $y$ position in the image and area, respectively.

To prevent rapid switching from one patch candidate selection to another, hysteresis is introduced. This forces the new candidate score to be larger than the current one by some margin. The hysteresis introduces a memory element into the candidate selection process as candidate scores are tracked over a period of time. It is therefore necessary to keep track of the detected magnetic patches IDs, and update them from frame to frame. 

\begin{figure} [!ht]
	\centering
	\includegraphics[trim=0 0 4cm 2cm, clip, width=0.6\textwidth]{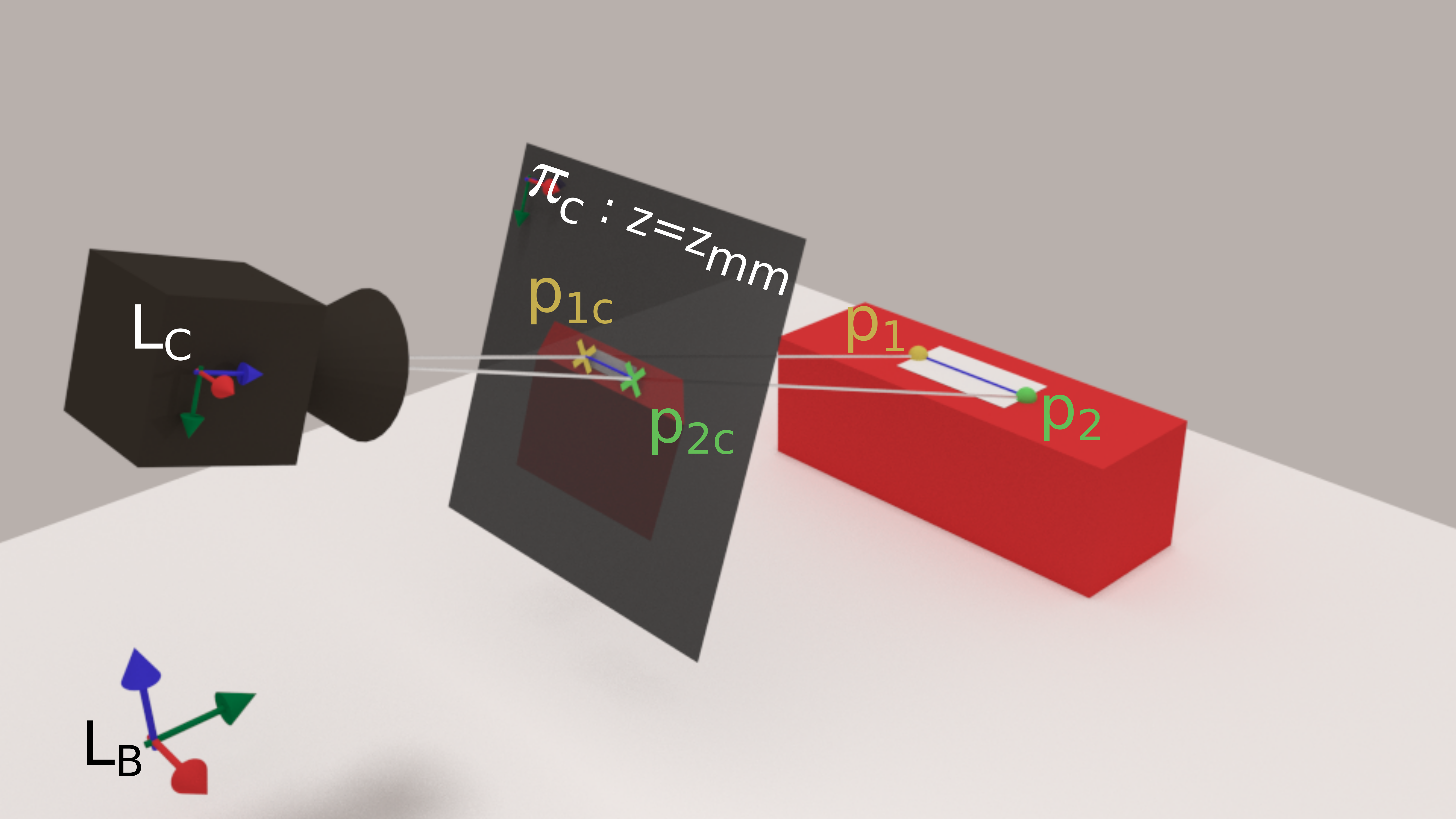}
	
	\caption{Endpoints of a patch in the image plane are transformed into the robot base frame $L_B$ using camera intrinsic parameters. The pose of the patch is calculated by intersecting the lines from the camera center through the rectangle endpoints in the image plane with the horizontal plane whose height is extracted from the 3D point cloud obtained from Realsense.}
	\label{fig:pnp_img_projekcija}
\end{figure}

The scoring scheme finally yields a single best patch candidate described with its corresponding rectangle. The rectangle's pose is determined by the pose of its major axis. In fact, only the endpoints of this axis in the RGB camera image are considered ($p_{1}$ and $p_{2}$ in Figure \ref{fig:pnp_img_projekcija}), which eliminates the effect of perspective distortion. As a first step in obtaining the rectangle's pose in the robot body frame $L_B$, the pixel positions $(x_{px},y_{px})$ of both endpoints' image projections are transformed into distance measure using camera intrinsic parameters, namely the focal length in pixels $z_{px}$ and millimeters  $z_{mm}$, along with the camera image resolution in pixels $x_{px} \times y_{px} $ using equation \ref{eq:proj1} for both $x$ and $y$ coordinate. The complete image plane projections of the endpoints $p_{1}$ and $p_{2}$ in the camera local frame $L_C$ are then obtained as $p_{1c}$ and $p_{2c}$, with the $z$ coordinate in the image plane $\pi_c : z = z_{mm}$.
\begin{equation}
x_{mm} = x_{px} \frac {z_{mm}}{z_{px}}
\label{eq:proj1}
\end{equation}

With a known position of the axis endpoints' projections in the local camera frame $L_C$, the same are obtained in the robot base frame, denoted $L_B$ in Fig. \ref{fig:pnp_img_projekcija}, using the direct kinematics transform matrix $T_{B}^{C}$. Camera position $p_{cam} |_{L_B} = (x_{cam}, y_{cam}, z_{cam})$ is also obtained in the robot base frame $L_B$ under the camera pin-hole model assumption, with the local camera position in the $L_C$ frame origin, using the same transform $T_B^C$. Now, lines passing from the camera origin through the projected endpoints $p_{1c}$ and $p_{2c}$ can be described in the x-z plane in the robot base frame with eq. \ref{eq:proj2}. Here, the index $i \in \{1,2\}$ denotes both axis endpoints $p_1$ and $p_2$. The equivalent linear equation holds for the y-z plane in the robot base frame. 
\begin{equation}
x = x_{cam} - \frac{x_{cam} - x_{p_ic}}{z_{cam} - z_{p_ic}} (z_{cam} - z)
\label{eq:proj2}
\end{equation}

What remains in determining the pose of the brick patch in $L_B$ is to find the $x$ and $y$ coordinates of $p_1$ and $p_2$ in $L_B$, when the patch height $z=patch\_height$ is substituted into the projection equations \ref{eq:proj2} for both $x$ and $y$ coordinates of $p_1$ and $p_2$. The patch height can be obtained from the organized point cloud generated from the Realsense depth image upon transformation into the camera and the base frame. Using the known location of the patch in the image, the height of the brick stack can be read in the organised point cloud. 

The pose of the wall pattern edge is obtained in a similar manner. This pose is detected once, transformed into the global map frame using the transformation matrix $T_{M}^{L}$ that contains the estimate of robots position and orientation in the map frame $L_M$, and reused for each subsequent brick drop. Similarly, the pose of the magnetic patch is also transformed into the global map frame using $T_{M}^{L}$ for calculation of the navigation goal pose during the $Alignment$ phase.

\section{Low-level control for navigation and manipulation}\label{sec:low_level_control}

The UGV is controlled in two ways: a) \textit{Map Navigation} uses Move Base motion planning to navigate to a desired goal in the 2D map while avoiding obstacles, b) \textit{Local Object Approach} controls the robot's forward and angular velocity directly without taking the obstacles into account.

\subsection{Map Navigation} 
\begin{figure} [!ht]
	\centering
	\includegraphics[width=1.0\textwidth]{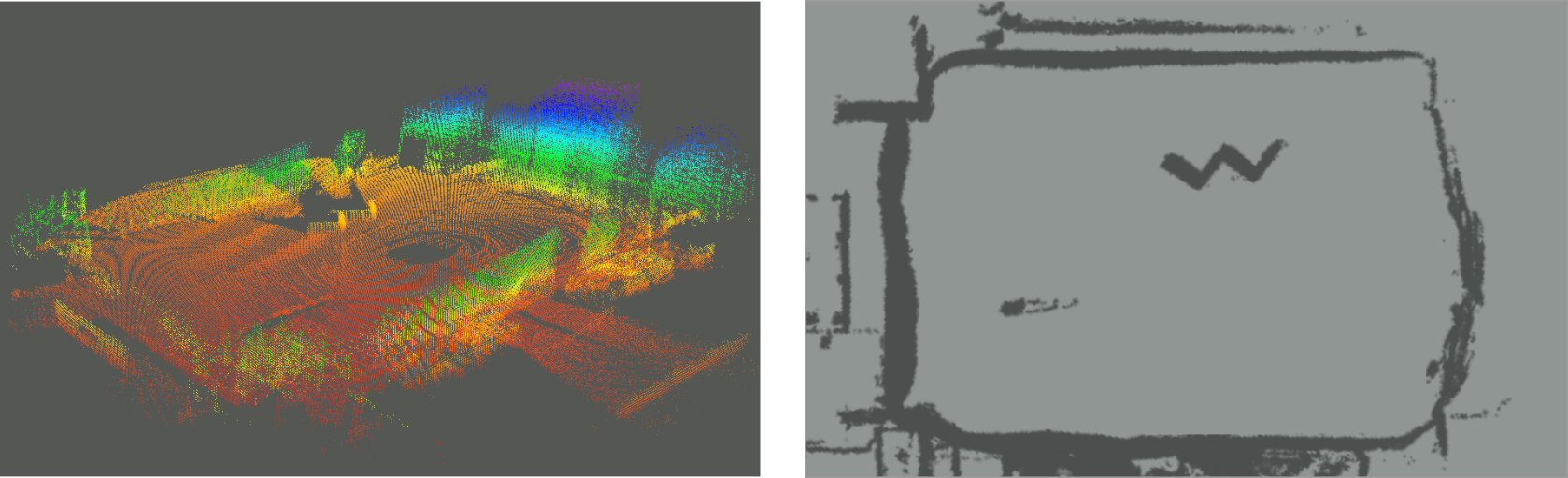}
	\caption{Point cloud of the Challenge 2 arena in Abu Dhabi, built by the Cartographer 3D SLAM algorithm is shown on the left. The 2D costmap generated from the 3D point cloud, used for the Move Base package, is shown on the right.}
	\label{fig:pointcloud_img}
\end{figure}

Google Cartographer SLAM \cite{Hess2016} is used to build a map of the unknown environment and localize the robot in it. 
Since the Move Base motion planning package needs a 2D costmap, it is generated using the 3D SLAM algorithm (Figure \ref{fig:pointcloud_img}) by filtering the Cartographer submap points based on their height. 
Points below the lower threshold are filtered out to avoid labeling ground as an obstacle. Similarly, points above the robot maximum height are removed as they do not present a real obstacle in platform navigation. 
Ground points that are far from the robot may be perceived as false positive obstacles if the ground is not completely flat, which can be solved by building "smaller" submaps with $30$ laser scans per submap.

Although the Husky A200 is a differential drive mobile robot and has the capability to rotate in place, in practice this motion causes significant vibration, particularly in cases when there is high friction between the rubber wheels and the terrain. These vibrations severely affect certain aspects of the system, most notably mapping and localization, as well as image processing algorithms. To mitigate this issue, the Timed-Elastic Band (TEB) planner for mobile robots is used as a local planner in the ROS navigation stack \cite{Rosmann2017}. The TEB planner uses online trajectory optimization to generate a local plan for the robot subject to a minimum-turning-radius constraint. The trajectory optimization used by the planner incorporates soft constraints on velocity limits, but it was found that the produced plan often does not satisfy these constraints. The algorithm was modified to constrain the UGV's forward and angular velocity, $v_x$ and $\omega_z$ respectively
\begin{equation}
\underline{v}_{x} \leq v_x \leq \overline{v}_{x}   
\end{equation}
\begin{equation}
|\omega_z| \leq \overline{\omega}_{z},
\end{equation}
while keeping the planned turning-radius
\begin{equation}
\frac{v_x}{\omega_z} = \frac{v_{x,planned}}{\omega_{z,planned}}.
\end{equation}

The initial trajectory generated by TEB satisfies the imposed turning-radius constraints by producing a motion plan with a high number of back and forth motion switches. In the original implementation, the initial trajectory is pursued and refined through iterative online optimization, reducing the number of forward-backward switches. In contrast to the original implementation, this work introduces initial conditions into the algorithm, suppressing the velocity commands until the iterative trajectory optimization yields a plan with a satisfactory low number of forward-backward switches. The initial conditions can also be generated by a timeout if the optimization gets stuck in a local optimum.

\subsection{Local Object Approach} 

\begin{figure} [!ht]
	\centering
	\includegraphics[width=0.9\textwidth]{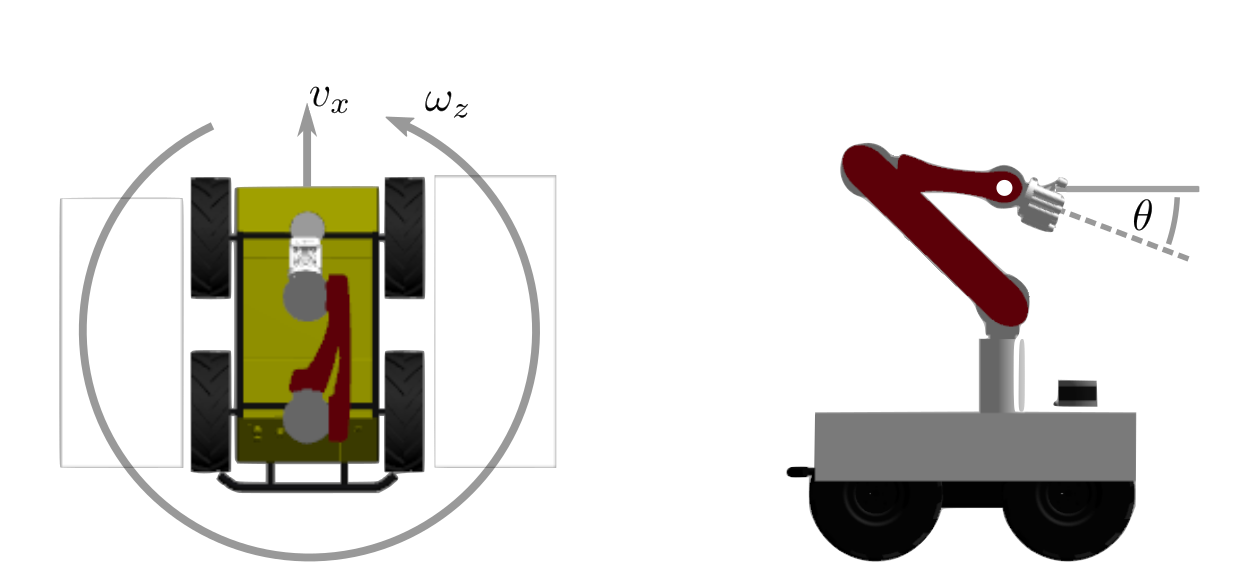}
	\caption{Control inputs used for the local approach: the top view of the mobile manipulator with its forward and angular velocity is shown on the left, and its side view with the camera pitch angle is shown on the right.}
	\label{fig:ctrl}
\end{figure}

Local Approach is used to track a desired object under the assumption that the object is in the camera field of view. During servo control, the control inputs that achieve this assumption are the UGV's forward velocity $v_x$, the UGV's angular velocity $\omega_z$, and the displacement of the camera pitch angle $\Delta{\theta}$ controlled by the robot manipulator (Figure \ref{fig:ctrl}). In the competition, a rough estimate of the initial brick stack position  was known a priori, and the mobile manipulator was navigated to a point in that area of the arena. Once there, the manipulator arm performed a rotational scanning motion searching for the stack of a particular color, if it was not already in the field of view. 

In a more general scenario, as originally announced in the MBZIRC challenge rules, the initial information about the arena, including the map and the positions of the brick stacks and the wall pattern, would be shared within the robot team. In this way, a quick search flight would be conducted by the UAVs in the robot team, providing the UGV with the necessary navigation information. Since the operation of the different types of robots was ultimately independent, a lawnmower search trajectory was planned for the UGV with the brick stack detection and localisation as objective. However, due to the time consumption of such a search pattern within a short experimental time slot, and thanks to the easily located brick stacks, the lawnmower search was not deployed. 

The position of the object in the image ($x_{img}$, $y_{img}$), and object distance $d_x$, measured in the robot frame $L_B$, are used as the feedback. Depending on the current state, $x_{img}$ and $y_{img}$ are acquired from either the brick stack, the magnetic patch, or the wall footprint detection. The object distance is obtained by filtering the Realsense camera point cloud with the detected object mask as the closest point in the UGVs forward direction. To deal with occasional poor distance measurements, a constant velocity model Kalman filter is used. Since the distance is measured with the RGBD camera, it is important to keep the desired object in the camera field of view during the approach.

Proportional controllers are used to approach the desired object, while keeping it in the center of the image:
\begin{equation}
v_x = - K_{d_x}^A (d_r - d_x)
\end{equation}
\begin{equation}
\omega_z = - K_{x_{img}}^A x_{img} 
\end{equation}
\begin{equation}
\Delta{\theta} = K_{y_{img}}^A y_{img} ,
\end{equation}
where $d_r$ is the desired object distance, $K_{d_x}^A$, $K_{x_{img}}^A$ and $K_{y_{img}}^A$ are proportional gains of the distance, $x$ and $y$ image position controllers respectively. 
\subsection{Visual Servo Brick Pickup}
Once the \textit{Two-Stage Approach} to the brick is finished, the reachability of the magnetic patch is checked, and if the patch is reachable, the \textit{Visual Servo Brick Pickup} algorithm is executed. This state is performed solely by the robot manipulator  while the UGV remains idle. The brick pickup motion is divided into four different stages: \textit{$x$ and $Pitch$ Visual Servo}, \textit{$y$ Visual Servo}, \textit{$yaw$ Visual Servo} and \textit{$z$ Approach}, all of which are executed in the robot frame $L_B$,  to decouple the controllers used for the different stages, simplifying the tuning process.  
\subsubsection{$x$ and $Pitch$ Visual Servo}
\begin{figure} [!ht]
	\centering
	\includegraphics[width=1.0\textwidth]{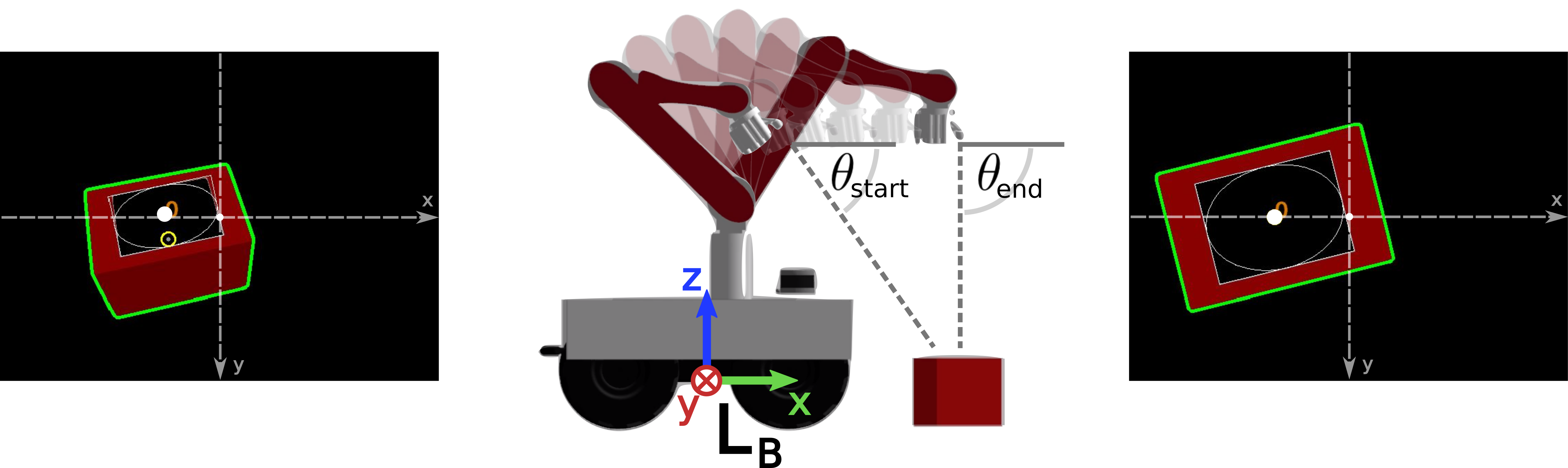}
	\caption{\textit{x and Pitch Visual Servo} stage of the \textit{Brick Pickup} behavior. The end effector movement direction is described with respect to the robot base frame $L_B$. Filtered camera images at the beginning and at the end of this stage are shown on the left and right, respectively. In both images, the position of the magnetic patch on the image $y$ axis is close to zero, the main difference being the camera pitch angle, denoted $\theta_{start}$ and $\theta_{end}$ in the middle image.}
	\label{fig:x_pitch_img}
\end{figure}

During the \textit{$x$ and $Pitch$ Visual Servo} stage of the visual servo algorithm, the $x$ position of the electromagnetic gripper and the camera pitch angle are controlled (Figure \ref{fig:x_pitch_img}). The goal of this stage is to orient the camera downward, while keeping the magnetic patch inside the camera image. Increasing the camera pitch angle causes the magnetic patch to move upward in the image. This is compensated for by moving the camera forward in the $x$ direction.

Two proportional controllers are active at the same time:
\begin{equation}
\Delta \theta = K_{\theta}^M (\theta_d - \theta)
\label{eq:xPitch1}
\end{equation}
\begin{equation}
\Delta x = -K_{y_{p}}^M y_{p},
\label{eq:xPitch2}
\end{equation}
where $\theta_d$ is the desired camera pitch angle, which is set to $\pi/2$ (camera facing downwards). $\Delta x$ represents the end effector displacement in the $x$ direction. 

This stage of the visual servo pickup is completed when the camera pitch angle reaches $\pi/2$ and the $y$ image coordinate of the magnetic patch $y_p$ is sufficiently close to zero (Figure \ref{fig:x_pitch_img}).

\subsubsection{$y$ and $yaw$ Visual Servo}

\begin{figure} [!ht]
	\centering
	\includegraphics[width=1.0\textwidth]{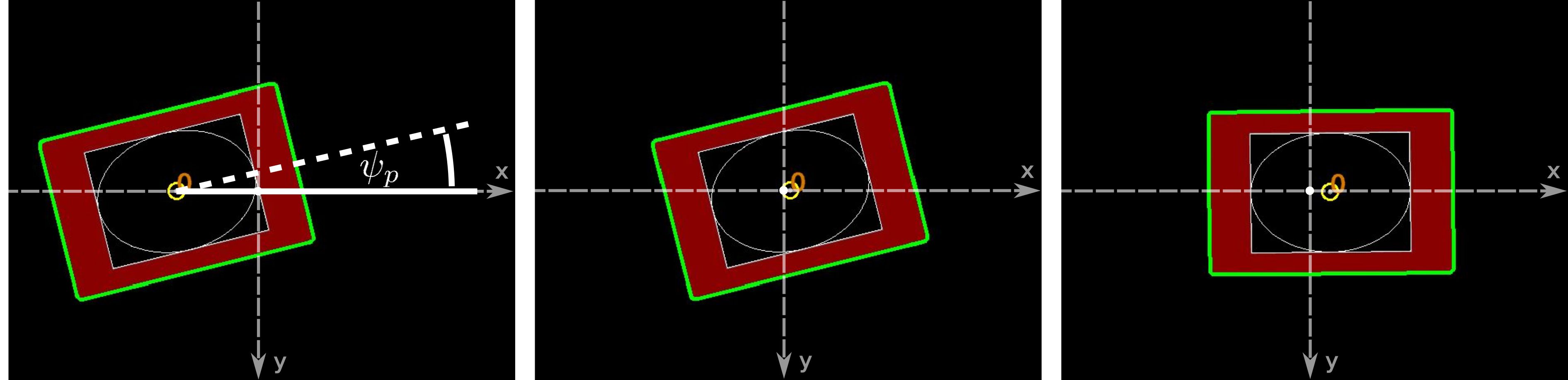}
	\caption{Filtered camera image after the \textit{x and Pitch Visual Servo} (left), \textit{y Visual Servo} (middle), and \textit{Yaw Visual Servo} (right). The sequence shows the progress of the pickup procedure. First, the end effector is positioned directly over the center of the magnetic patch, then the effector is aligned with the axes of the patch to ensure proper orientation of the bricks. The final offset of the camera from the center of the patch accounts for the displacement of the camera with respect to the center of the end effector.}
	\label{fig:y_yaw_img}
\end{figure}

Once \textit{$x$ and $Pitch$ Visual Servo} is finished, the remaining $x_{p}$ and orientation errors are eliminated with the \textit{$y$ Visual Servo} and \textit{$yaw$ Visual Servo} proportional controllers. The following proportional controllers are designed:
\begin{equation}
\Delta y = - K_{x_{p}}^M x_{p}
\label{eq:yServo}
\end{equation}
\begin{equation}
\Delta \psi = - K_{\psi_{p}}^M \psi_{p},
\label{eq:yawServo}
\end{equation}
where $\psi_{p}$ is the magnetic patch camera orientation, and $\Delta\psi$ is the end effector yaw displacement
(Figure \ref{fig:y_yaw_img}). 
\subsubsection{$z$ Approach}
\textit{$z$ Approach} starts once the magnetic patch is sufficiently close to the image center, with a sufficiently small $\psi_{p}$. At this stage, the center of the magnetic patch is located directly bellow the RGB camera, so the electromagnetic gripper is displaced to account for the offset between the camera and gripper centers.

Distance measurements from the Realsense camera's depth module are used to estimate the distance between the magnetic patch and the electromagnetic gripper $d_z$. Knowing that the height of the magnetic patch in the robot base frame $L_B$ is $N$ times the height of a single brick $h_b$, where $N$ is a positive integer, the final value of measurement $d_z$ is set to the closest $Nh_b$ value.  
The following proportional controller
\begin{equation}
\Delta z = K_{d_{z}}^M d_z
\end{equation}
is designed for the \textit{$z$ Approach} and finishes at the first triggering of the contact sensor. 
\section{Results} \label{sec:experimental}

Discussion of the performance in the competition is given in section \ref{comp_perf} and presents results in an outdoor scenario under realistic, real-world conditions.
We conducted additional, post-competition experiments to evaluate some of the previously described algorithms in an indoor environment, using the Optitrack motion capture system to provide ground truth measurements. 

\subsection{Competition performance} \label{comp_perf}
The first real performance test for our system was the competition. The Challenge 2 environment was designed as a single arena with two separate wall-building sites, one for the UAVs and one for the UGV. Each of the two walls was to be built using dedicated bricks. The bricks designated for the UGV were set up as structured brick stacks, separated by color. 

Each team had three $25$ minute runs to attempt the Challenge, either in autonomous or manual mode. The location of the brick stacks and the wall footprint changed in each run, but the layout of the Challenge was similar: the brick stacks were positioned on the side of the arena closest to the UGV and were visible from the starting position (especially since the starting orientation of the UGV could be set arbitrarily by the team) while the building site was somewhere on the other side of the arena, narrowing the area for search. 

In the three runs of the Challenge, the UGV managed to autonomously pick up multiple bricks and place them in the baskets. In the second run a green brick was placed on the wall footprint fully autonomously by the UGV, giving the team a score of $0.44444$ points, which was enough for 5th place in the Challenge 2 rankings. 

It was difficult to collect data in the competition, and, consequently, evaluate performance. However, certain observational conclusions were reached. No significant problems were noticed regarding the navigation and mapping algorithms. The two-dimensional maps built by Cartographer seemed to correspond to the environment, without any evident flaws. An example of a Challenge 2 arena map built during the competition is shown in Figure \ref{fig:pointcloud_img}. The TEB planner managed to produce and execute a feasible plan in every instance. The planned trajectories satisfied the minimum-turning-radius constraint and thus successfully resolved a previously described base-vibration problem. The only evident drawback of move-base navigation is time-consumption, discussed in the "Limitations" section \ref{sec:limitations}.  

Variable lighting conditions caused detection problems that negatively affected visual-feedback dependent algorithms. This issue was less pronounced during local approach and more so in manipulation tasks. Apart from that, when detection performed well, there were no problems with the manipulation.

\subsection{Post-competition evaluation}
To further test overall system performance, we conducted additional post-competition experiments, using the Optitrack system as a benchmark to provide ground truth data. In the first set of experiments, the goal was to evaluate the performance of the \textit{Brick Loading} process with two measures: error in brick-pose estimation and brick-loading success rate. 

The second set of experiments focused on evaluating the \textit{Brick Unloading} performance with similar measures: differences in estimated wall pose compared to ground truth data and success rate of the algorithm. For both of the aforementioned experiment sets, ground truth data was obtained by placing Optitrack fiducials on the Husky, the bricks, and the wall pattern. 

In the final experiment, the goal was to evaluate the autonomy of the system performing the entire Challenge as well as the time required to complete the mission.

\subsection{Brick loading experiment}

In this experimental scenario the robot is assumed to have located the brick stack. To complete the brick loading task, the robot must identify an individual brick of a given color, estimate its pose, navigate to it and pick it up. All of these actions are performed fully autonomously. To evaluate the brick pose-estimation, the brick and the UGV poses are recorded via Optitrack at the beginning of the \textit{Alignment} phase of the two-stage approach, and compared to the pose obtained using the algorithm described in section \ref{sec:detection}. The results of each experiment are shown in Table \ref{tab:tab_1}.

The experiment was repeated seven times, with a $100\%$ success rate. Table \ref{tab:tab_1} shows that the developed algorithm can cope with various initial relative orientations between the UGV and the brick. 
Since the estimated brick pose is used only to navigate to a starting point for the \textit{Final Approach} visual servo, the system is robust enough to successfully pick up the brick even if the distance to the brick is underestimated. 

We determined experimentally that the configuration of the platform and the robot arm allows for a perpendicular approach error of approximately $26^\circ$. All observed errors fall comfortably within this range.

\begin{table}[!t]
	\centering
	\caption{Brick approach and loading experiment showing differences between the true (relative to the robot, measured in the Optitrack frame) and the estimated object pose (relative to the robot, measured in the robot frame $L_B$) for each of the experiments and the mean absolute error (MAE) across all experiments. Since a success rate of $100\%$ indicates that the position of the brick is correctly identified, we show only the total distance of the brick from the Husky and error in estimating this distance.}
	\begin{tabular}{|c|c|c|c|c|c|}
		\hline
		Run no. & \makecell{Patch \\ distance [$m$]} & \makecell{Patch \\ distance error [$m$]} & \makecell{Patch \\ orientation [$^\circ$]} & \makecell{Patch orientation \\ error [$^\circ$]} & \makecell{Success} \\ \hline
		1 &  1.632 & -0.153 & -87.72  &  -3.00 & Yes  \\ 
		2 & 1.673 & -0.173 &  -105.37 &  -5.43 & Yes  \\ 
		3 & 1.670 & -0.144 &  -138.48 &  -6.23 & Yes 	\\ 
		4 & 1.702 & -0.159 &  -161.46 &  -7.96 & Yes 	\\ 
		5 & 1.761 & -0.249 &  -66.69  &  -2.89 & Yes 	\\ 
		6 & 1.625 & -0.143 & -89.38  &  -6.47 & Yes 	\\ 
		7 & 1.677 & -0.144 & -52.12  &  6.90  & Yes 	\\ \hline
		\multicolumn{2}{c}{} & \multicolumn{1}{|c|}{MAE[m] = 0.166} & \multicolumn{1}{|c|}{} & \multicolumn{1}{|c|}{MAE[$^\circ$]=5.5} & \multicolumn{1}{c}{}
		\\
		\cline{3-3}\cline{5-5}
		
	\end{tabular}
	\label{tab:tab_1}
\end{table}

\subsection{Brick unloading experiment}

The effectiveness of the brick unloading algorithm was evaluated in a similar manner. The UGV is initially placed at an arbitrary distance from the wall footprint, with an arbitrary relative orientation between the two. The robot must position itself perpendicular to the footprint's long axis and at a predefined point relative to the rightmost point of the pattern. The scenario's final step places a brick on the wall, an action that comprises of picking the brick up from the basket and unloading it in an appropriate location on the template or wall.  The results of each run and some summary statistics are shown in Table \ref{tab:tab_2}.

The experiment was performed ten times with a success rate of $100\%$. The outcome of the experiment is considered successful only if at least half of the brick is placed inside the wall pattern. With respect to the estimation of distance to the pattern, the system behaves similarly to brick loading, since the final step is a visual servo to the patch that keeps the brick centered on the pattern across its shorter axis. However, since the starting point for this servo procedure is constructed based on the first pattern servo goal (which is the rightmost point of the pattern), the error in brick placement  along the longer axis of the pattern is not accounted for, generating gaps between bricks. As can be seen in the following section, the gaps between bricks do not increase over time, suggesting that the longitudinal errors in brick placement are, for the most part, generated by the TEB planner's goal tolerance and, to a lesser degree, by errors or drift in localization.

\begin{table}[!t]
	\centering
	\caption{Brick unloading experiment showing the effectiveness of the brick unloading procedure. The errors in orientation of the pattern are measured by comparing the orientation of the placed brick to the orientation of the pattern, measured in the Optitrack frame. The distance and the orientation of the pattern with respect to the robot when it is first detected is also shown.}
	\begin{tabular}{|c|c|c|c|c|}
		\hline
		Run no. & \makecell{Pattern \\ distance [$m$]} & \makecell{Pattern \\ orientation [$^\circ$]} & \makecell{Brick placement \\ orientation error [$^\circ$]} & \makecell{Success} \\ \hline
		1 & 1.232 &  2.79  &  4.96 & Yes  \\ 
		2 & 0.934 &  52.22 &  1.41 & Yes  \\ 
		3 & 1.208 &  1.03 &  1.03 & Yes \\ 
		4 & 1.263 &  28.05 &  4.91 & Yes 	\\ 
		5 & 1.125 &  3.74  &  4.91 & Yes 	\\ 
		6 & 1.269 &  0.57  &  6.85 & Yes 	\\ 
		7 & 1.211 &  6.65  &  9.55  & Yes 	\\
		8 & 1.321 &  1.68 &  9.88  & Yes 	\\
		9 & 1.457 &  2.55  &  2.38  & Yes 	\\
		10 & 1.115 & 2.73  &  4.31  & Yes 	\\
		\hline
		\multicolumn{3}{c}{} & \multicolumn{1}{|c|}{MAE[$^\circ$]=5.02} & \multicolumn{1}{c}{}
		\\
		\cline{4-4}
		
	\end{tabular}
	\label{tab:tab_2}
\end{table}

\subsection{Wall Building Scenario}

\begin{figure} [!ht]
	\centering
	\includegraphics[width=0.95\textwidth]{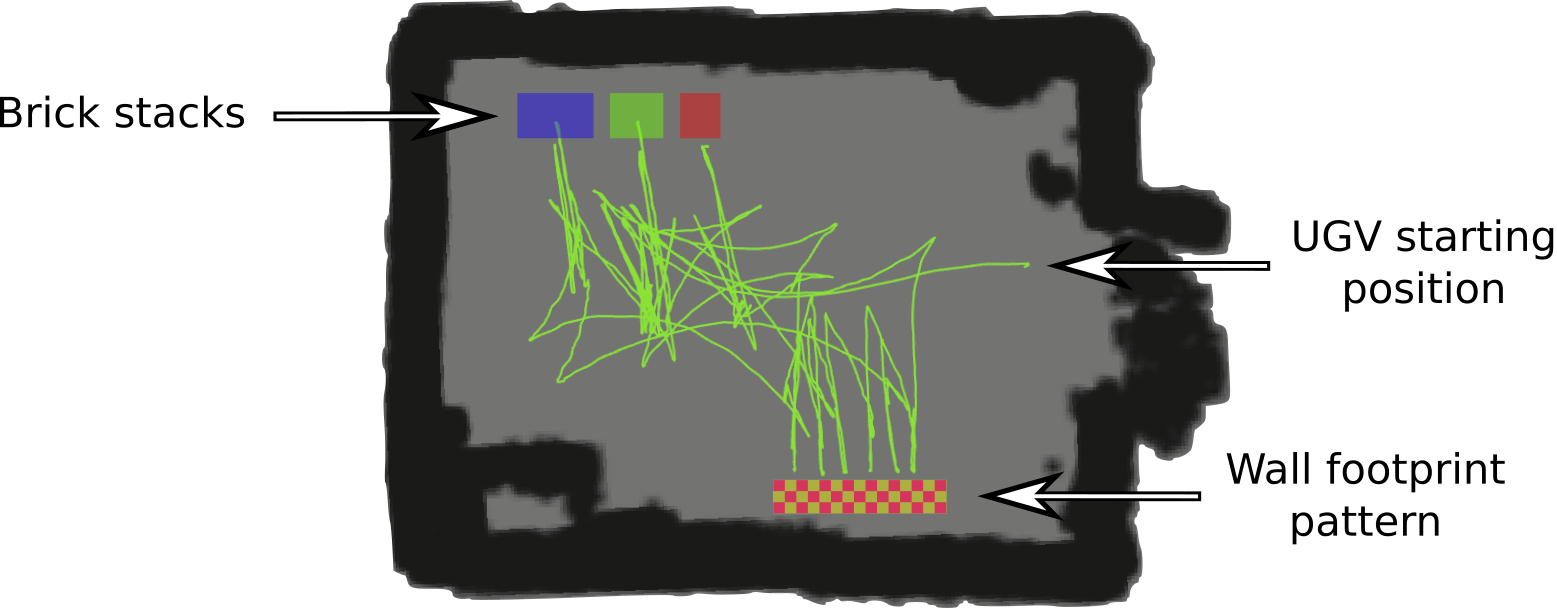}
	\caption{Path traversed during the fully autonomous wall building experiment. The wall footprint pattern, brick stack position, and the UGV starting position are shown in a 2D map built during the experiment. Dimensions of the arena are $10\times7.5m$. Distance between the UGV start position and the brick stacks is $7m$. The distance between the starting position and the wall footprint is $3.5m$.}
	\label{fig:full_path}
\end{figure}

An experiment to evaluate  the entire autonomous wall building scenario was performed in the Challenge mock-up environment built on the University campus, shown in Figure \ref{fig:husky_bricks}, with the experiment layout as shown in Figure \ref{fig:full_path}. The experiment layout is inspired by the one in the Challenge. The bricks are set up as structured brick stacks separated by color, and the wall pattern is placed on the other side of the arena. Similarly to the competition, there are no obstacles between the brick stacks and the building site.

\begin{figure} [!ht]
	\centering
	\includegraphics[width=1\textwidth]{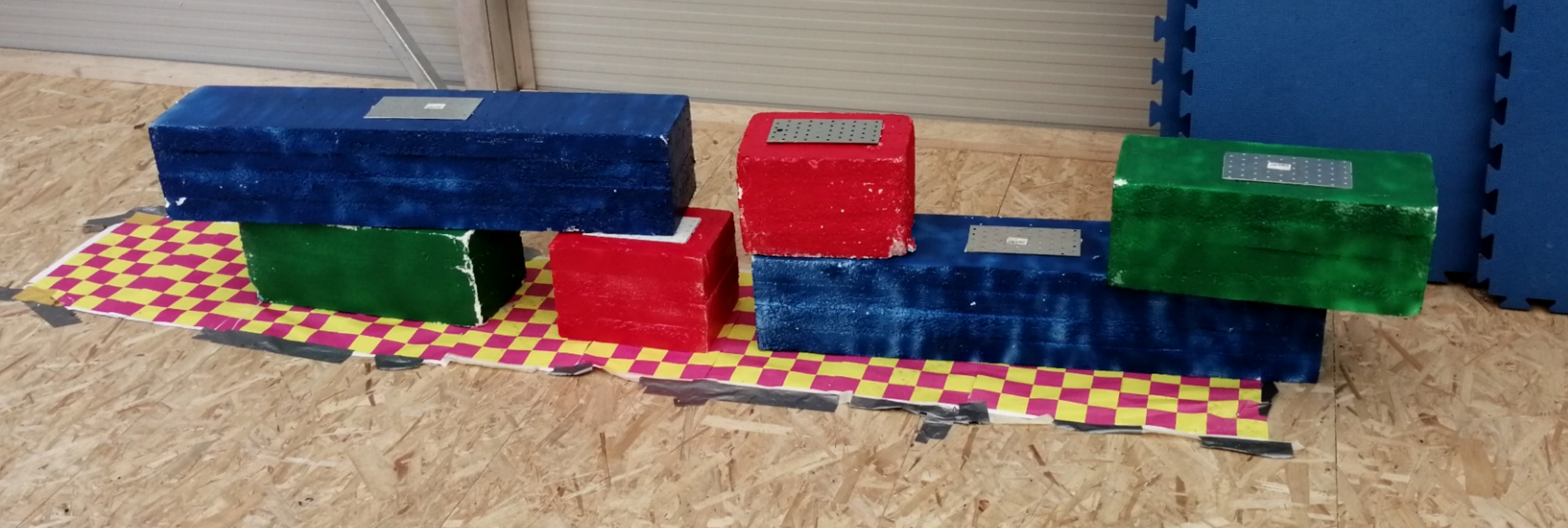}
	\caption{Six brick wall built in an autonomous full-scale MZBIRC 2020 Challenge 2 experiment. Total of three trips between brick stacks and wall were required and all bricks were placed within the boundaries as proposed by the competition.}
	\label{fig:wall}
\end{figure}

The system is provided with a desired wall configuration, along with the approximate initial position of the brick stacks, and that of the wall footprint. The wall building speed highly depends on the desired wall structure, as different colored bricks take up different amounts of space in the brick baskets. This situation makes some wall structures less desirable speed-wise, as they require multiple trips between the brick stacks and the wall footprint to complete the wall. Another factor that affects the overall speed of the system is the Challenge layout: the initial brick stacks position, the initial UGV position, and the wall-footprint position. Performing full runs of the experiment is time consuming, but from the several runs that were completed, it was observed that the UGV is capable of transporting and laying 6 to 10 bricks within $25$ minutes \footnote{Time limit for the Challenge}, depending on the wall configuration. The final result of such a run is shown in Figure \ref{fig:wall}. 

The wall, consisting of two blue, two green, and two red bricks was successfully built without requiring any type of manual intervention. It took the robot $23$ minutes and $37$ seconds to perform the entire experiment, and all  bricks were placed with sufficient precision to satisfy Challenge rules. A video of the experiment is available at \url{https://youtu.be/0C_E7tGCZ5k}. 

Overall, the experiments shown in this paper validate our approach and confirm that precision of the constrols, 3D-map localization, navigation and the pose-estimation from RGB images satisfy the Challenge requirements. Finally, our custom-developed, passively-compliant, electromagnetic gripper successfully addressed the end-effector misalignment problem in every brick loading and unloading procedure. 

\section{Conclusion}\label{sec:conclusion}

This paper presented the LARICS team's approach to the Challenge 2 of MBZIRC 2020 competition, as well as the post-competition experimental setup and results. Both the brick loading and unloading experiments were performed with a $100\%$ success rate, the latter of which resulted in brick placement sufficiently precise to satisfy Challenge rules. 
During the full Challenge experiment, a wall consisting of two red, two green and two blue bricks, shown in Figure \ref{fig:wall}, was built autonomously. The entire trial lasted $23$ minutes and $37$ seconds. 

\subsection{Limitations}\label{sec:limitations}

Our approach described in this work is specifically designed for a semi-structured scenario, as described by the Challenge 2 rules, and as such does not generalize to other non-structured scenarios.  

The only obstacles in the competition arena were the brick stacks and an UAV wall-building site (as seen in Figure \ref{fig:pointcloud_img}). Even though the TEB planner takes obstacles into account, a more crowded environment could introduce obstacle-avoidance problems. Since the planner is configured to use a large minimal-turning-radius, in environments with many obstacles it may be unable to produce a feasible plan due to numerous constraints. This conflict could be tackled by reducing the minimum-turning-radius or using a different planner. 

In the Challenge, the bricks were initially structured in stacks separated by color, which organization was exploited in the local approach and manipulation algorithms. The local approach does not take obstacles into account, as long as they do not interfere with the visual feedback. This constraint could present a problem in a more general scenario, if there are obstacles in close proximity to the target object. Likewise, the manipulation algorithms were designed with a structured brick stack in mind and would require modifications in the case of a chaotic brick pile.

Object detection based on HSV filtering presents an obvious general-case limitation in terms of object and environment texture. The bricks in the competition were color-coded to stand out from the environment. Nevertheless, detection proved to be difficult in the Challenge 2 scenario. Each trial was conducted under different sunlight conditions, forcing the adjustment of HSV thresholds used for both brick and wall-footprint detection. Brick-stack and wall-footprint detection, which are solely color-based, performed well even with a less than optimal HSV thresholds, while grasp-point patch-detection, which is based on both color and shape, was more error-prone, especially in cases where the manipulator creates a moving shadow on the brick.

In our post-Challenge tests, bricks were placed on the wall-footprint with an orientation MAE of $5.02^\circ$, and a more significant position error as evidenced by gaps between bricks. These errors were more pronounced in a full challenge scenario, due to the fact that the UGV does not detect the wall pattern's pose at each brick drop, but memorizes its map pose and reuses it. This issue could be avoided using fresh detection of the previously placed bricks or the wall pattern edge during the alignment, but was not found to be necessary.

There is room for improvement in speed. Most of the time is spent in the navigation stages of the Challenge, mainly in the stages that involve Move Base motion planning. This situation is due to the TEB planner re-planning until the goal pose-tolerance is satisfied. Since the trajectory produced by the TEB planner is constrained by a minimum-turning-radius, a small error in the achieved pose requires a non-trivial maneuver to eliminate it. Furthermore, we suspect that the manipulation speed could be improved using not only the image position of the detected magnetic patch, but also its estimated global pose to plan end-effector motion.

\subsubsection*{Acknowledgments}
Research work presented in this article has been supported in part by Khalifa University of Science and Technology, Abu Dhabi, UAE, by European Commission Horizon 2020 Programme through project under G. A. number 810321, named Twinning coordination action for spreading excellence in Aerial Robotics - AeRoTwin, and project Heterogeneous autonomous robotic system in viticulture and mariculture (HEKTOR) financed by the European Union through the
European Regional Development Fund-The Competitiveness and Cohesion Operational Programme (KK.01.1.1.04.0041). Work of Frano Petric was supported by the European Regional Development Fund under the grant KK.01.1.1.01.0009 (DATACROSS).

\bibliographystyle{apalike}


\end{document}